\documentclass{article} 
\usepackage{iclr2022_conference}

\usepackage{hyperref}
\usepackage{url}
\usepackage{graphicx}
\usepackage{tikz}
\usepackage{tikz-cd}
\usepackage{booktabs}
\usepackage{multirow}
\usepackage{cleveref}
\usepackage{listings}
\usepackage{csquotes}
\graphicspath{ {./figures/} }

\usepackage[lofdepth,lotdepth]{subfig}

\title{Cut the CARP: Fishing for zero-shot story evaluation}

\newcommand*\samethanks[1][\value{footnote}]{\footnotemark[#1]}

\author{Shahbuland Matiana\thanks{Equal contribution.}, JR Smith\samethanks, Ryan Teehan\samethanks$\,\,$\thanks{Also, Charles River Analytics}, Louis Castricato\samethanks[1], Stella Biderman\samethanks[1], \\
\textbf{Leo Gao, Spencer Frazier}\\
EleutherAI\\
Mail to: \texttt{lcastric@gatech.edu}
}

\newcommand\sysname[1]{{\sc carp}}

\iclrfinalcopy 
\begin{document}

\maketitle

\begin{abstract}

Recent advances in large-scale language models \citep{t5,gpt3} have brought significant qualitative and quantitative improvements in machine-driven text generation. Despite this, generation and evaluation of machine-generated narrative text remains a challenging problem. Objective evaluation of computationally-generated stories may be prohibitively expensive, require meticulously annotated datasets, or may not adequately measure the logical coherence of a generated story's narratological structure.

Informed by recent advances in contrastive learning \citep{radford2021learning}, we present Contrastive Authoring and Reviewing Pairing (\sysname{}): a scalable, efficient method for performing qualitatively superior, zero-shot evaluation of stories. We show a strong correlation between human evaluation of stories and those of \sysname{}. Model outputs more significantly correlate with corresponding human input than those language-model based methods which utilize finetuning or prompt engineering approaches. We also present and analyze the Story-Critique Dataset, a new corpora composed of 1.3 million aligned story-critique pairs derived from over 80,000 stories. We expect this corpus to be of interest to NLP researchers.
\end{abstract}

\section{Introduction}

Recent breakthroughs in natural language processing (NLP) and natural language generation (NLG) have revitalized interest in applying computational methods to story  \citep{see2019massively,xu2020megatroncntrl,nichols2020collaborative,fang2021transformerbased,hazarika2021zeroshot}. Automated Story Generation is the challenge of designing an artificial intelligence system that can generate a story from a minimal number of inputs---often a simple prompt and some storytelling primitives. Even with modern deep learning techniques this is a significant challenge, as people expect stories to be \emph{consistent} and \emph{coherent}, two things that transformers are not particularly good at doing across long passages \citep{gpt3,yao2019plan}.

Another reason automated story generation is challenging is that automated story \emph{evaluation} is challenging. In other domains such as image generation \citep{karras2019style, patashnik2021styleclip, galanos2021affectgan}, strategic game-playing \citep{alphazero, alphago}, and planning~\citep{karkus2017qmdp,fan2019automatic}, powerful models for evaluating objects of interest have lead to more powerful models for generating them. Unfortunately, methods for automatically evaluating stories, such as ROUGE \citep{lin2004rouge}, are extremely limited in their ability to accurately assess models and are easily Goodharted \citep{belz2006comparing,cohan2016revisiting,schluter2017limits,schluter2016approximate,eyal2019question}. Recently, researchers interested in story evaluation have developed more sophisticated techniques for evaluating stories \citep{akoury2020storium,dou2021scarecrow} as well as the Purdy Index \citep{purdy2018predicting} and the Fabula-Entropy Index \citep{castricato2021formal,castricato2021fabula}. Unfortunately, all of these methodologies rely on costly human evaluation and are therefore not suitable for integrating into end-to-end pipelines.

In this paper, we introduce CARP: a transformer-based model for automated story evaluation. CARP is a contrastive model, analogous in design to CLIP \citep{radford2021learning}, that learns to align the embedding space of two distinct encoder models, one that processes a story and one that processes feedback on that story. The result is that a passage's embedding has a higher similarity to those of its critical reviews. This way, CARP is capable of zero-shot story classification and review ranking, and is able to score stories based on their similarity to selected reviews.

\begin{figure}
    \centering
    \includegraphics[width=0.8\textwidth]{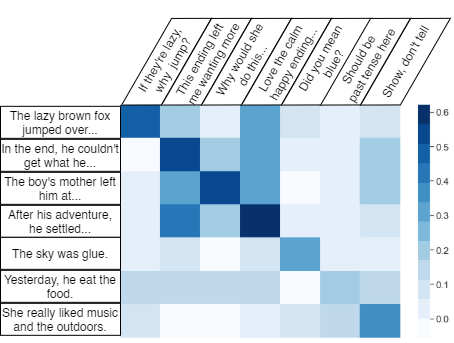}
    \caption{CARP learns to align representations of passages and corresponding representations of critiques.}
    \label{fig:cover}
\end{figure}

\subsection{Representations of Knowledge in Computational Narratology}

The actual text of a story does not fully encompass the space of all true statements about the storyworld. As a result, a crucial aspect of modeling criticism is handling the representations of the story as possessed by the author and the reader, respectively. In the early storytelling literature this was circumvented by using procedural graphical representations of the storyworld, effectively building into the model a storyboard it can reference. While such representations can be very useful, they have limited expressivity, are difficult to make for book-length stories, and do not capture much nuance. Some modern models, like CAST \citep{penginferring}, which aim to emulate a reader by creating a graphical representation of the story, suffer from similar issues.

End-to-end methods are typically more expressive than graphical methods of knowledge representation. However, due to the lack of constraints typically placed on their representations, they often go off-topic or focus on irrelevant details \citep{yao2019plan,dou2021scarecrow}. In general, the representations of autoregressive language models are not very robust \citep{tamkin2021understanding}, and language models often generate self-contradictory or nonsensical stories \citep{yao2019plan}.

To address this issue, we propose using constrastive learning for grounded language modeling \citep{fleischman2005verbs}. Methods like CLIP \citep{radford2021learning} are significantly more robust than naive classifiers, which we hope will allow CARP to learn more accurate and consistent evaluation of stories.

\subsection{Contrastive Learning with Transformers}

The core idea of contrastive learning is to learn to produce embeddings matching the correct target and to avoid embeddings that match other incorrect targets. Contrastive training first takes aligned pairs of data, and produces a pair of embeddings for each using encoder networks. Then, during training, a batch of $n$ of these pairs is taken and both embedding networks are trained to produce unit-length normalized embeddings which are as similar to each other within each pair as possible, and as dissimilar to the other embeddings in the batch as possible. In other words, the network attempts to make the similarity matrix as close to a diagonal matrix as possible.

Our work is largely inspired by CLIP by \cite{radford2021learning} which uses two such encoders; one encoder is used for images and another for the captions associated with those images. Contrastive learning between these aligned pairs produces both an image encoder and a text encoder. The similarity is measured using dot product in CLIP. The loss is a combination of the cross entropy classification losses for picking the right text given an image and vice versa. 

\subsection{Our Contribution}

Our paper makes several contributions to both the theory and practice of Computational Narratology.

\begin{itemize}
    \item We introduce CARP, the first automated model for story evaluation that can be incorporated into end-to-end deep learning pipelines.
    \item We show that contrastive learning is an effective tool for learning 
    \item We introduce the Story-Critique dataset, which we believe will be useful to narratology and language modeling researchers.
\end{itemize}

\subsection{Outline}

The rest of the paper is organized as follows: in \Cref{sec:dataset} we introduce and explore the Story-Critique Dataset that we will use to train our model; in \Cref{sec:model} we detail our model's construction and implementation; in \Cref{sec:evaluation} we discuss our evaluation methodology; in \Cref{sec:results} we analyze the results; and finally in \Cref{sec:conclusion} we draw conclusions and discuss future work.

\section{The Story-Critique Dataset}

In order to train our model, we needed a dataset of pairs (text, critique). Despite the widespread availability of textual datasets, we were unable to find any that included inline critiques. Therefore, we decided to create the Story-Critique Dataset. Like BookCorpus \citep{Zhu_2015_ICCV} and BookCorpus2 \citep{gao2020pile}, the Story-Critique Dataset consists of stories that have not been formally published. After speaking with the owners and hosts of the data we wished to use, we identified three major concerns about data use:
\begin{enumerate}
    \item The data authors do not wish for the full text of their stories to be made public.
    \item The data authors do not wish for their stories to be traced back to them as individuals.
    \item The data hosts do not wish to become publicly known.
\end{enumerate}

To address these concerns, we took several steps to preserve anonymity and privacy of both the data authors and the data hosts described in the subsequent sections. Given the sensitive nature of the dataset, it will be only made available to researchers for purposes covered by fair use and in the preprocessed and anonymized state described below. Additionally, we have decided to take the unusual step of relinquishing control of the dataset itself, leaving control of the dataset in the hands of the hosts with whom the data authors originally entrusted\footnote{A prior version of this paper stated that we would make the anonymized data publicly available. After further discussion with the data hosts they will be the sole determiners of if or when the data is made available}.

\subsection{Dataset Structure}\label{sec:dataset}

The dataset consists of more than 80,000 unique stories with 1,378,696 total critiques. Every critique refers to a specific passage of the story, and so we construct 1,378,696 (passage, critique) pairs for training. Passages are not indexed by story IDs to make reconstructing the stories more difficult. All data entries are anonymized --- unique identifiers including comment ID, submission IDs, URLs, and proper nouns have been removed. The critique type and word count are retained.

\subsection{Preprocessing} 

Preprocessing consists of three steps. First, splitting the story text into chunks corresponding to each inline critique. Second, masking sections of the critique directly quoting sections of the story. This is key to avoiding overfitting due to data leakage, otherwise the model could identify (passage, critique) pairs by matching the quotes \citep{biderman2020pitfalls,elangovan2021}. Third, anonymizing the data by removing proper nouns. This last step both preserves authorial privacy and prevents another form of data leakage in which the model matches stories with critiques based on overlapping proper noun preferences in said story.

\subsubsection{Quote Masking}

Removing story quotes from critiques is important for our model's design, but a simple matching algorithm risks removing common n-grams that should otherwise be preserved. It is also not sufficient to use regex matching to identify text that appears between quotation marks, as this can miss some direct quotes and erroneously identifies some revisions to the original passage as quotes. Instead, we use an approach based on the Longest Common Subsequence (LCS). By treating each word in the review and passage as a ``character'' --- after removing capitalization and punctuation --- we identify the LCS as the phrase to be masked in the original review (while retaining punctuation and capitalization). After some experimentation, we decided on a threshold size of 4 words for LCS quote identification, which successfully avoids misidentification of common phrases. Once an LCS of a size greater than or equal to 4 is identified, we mask each instance of the quote in the review with \texttt{[quote]} and pass the passage-critique pair for further processing.

\subsubsection{Anonymization}

Proper nouns --- both within the story and without --- can contribute to data leakage in collected reviews and compromise the anonymity of the authors. We use the BERT Named Entity Recognition model \citep{tjong-kim-sang-de-meulder-2003-introduction} and the \texttt{neuralcoref} package for coreference resolution\footnote{https://github.com/huggingface/neuralcoref} to identify names and other proper nouns. Once these names are identified, they are replaced with generic names indexed by numbers (ex. John0, Sam1, etc.). Manual spot-checks of 200 replacements revealed no instances of missed names. The anonymized and masked passages and reviews are then used to train the model.

\subsubsection{Final Processing}

After masking and anonymization, we found that a small fraction of reviews were short and unhelpful (i.e. empty strings, ``lol'', ``good'', ``haha'').  We removed all passage-review pairs with passages or reviews containing less than 8 characters (75,000 pairs or around 5.5\% of the total dataset). Finally, remaining passages are tail truncated at the token level, independent of sentences, or tail padded to fit a uniform context length of 512 tokens for the model.

\subsection{Analysis of Dataset Quality}

A common issue with human criticism datasets is that much of it is uninformative. People --- especially people posting on the internet where most ML datasets come from --- will frequently write brief, insulting, or otherwise unhelpful responses as ``feedback.'' In order to assess the extent to which this is a problem for the Story-Critique (SC) dataset, we measured the length (in word count), sentiment, and toxicity of reviews. For sentiment, we used BERT \citep{devlin2018bert} fine-tuned on sentiment analysis to classify responses as positive or negative. To identify toxicity, we used detoxify\footnote{https://github.com/unitaryai/detoxify} \citep{hanu2020detox}, a library containing multiple models that can score passages of text for various features considered harmful. Specifically, we used it to measure how insulting or toxic responses with negative sentiment were. We use the Writing Prompts (WP) dataset \citep{fan2018hierarchical} as a baseline, as we judge it to be the existing dataset that is most similar to ours. While its main use is for prompts paired with stories, the WP dataset also contains direct responses to its stories. We compare various aspects of these responses to those in our SC dataset.

\begin{figure*}[ht]
{}\centering
    {\includegraphics[width=0.5\linewidth]{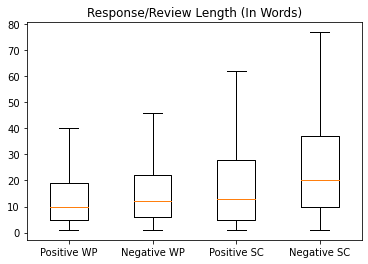}}
\caption{Box plots of story lengths for the Writing Prompts (WP) and Story-Critique (SC) datasets, broken down by sentiment. All four groups exclude outliers from in the plot.}
\label{fig:data}
\end{figure*}

Generally, we found that WP responses were shorter and on average more positive when compared to the SC responses. 67\% of the responses in WP were identified as containing a positive sentiment, while only 32\% of SC were.  On average, positive responses in both datasets were very short (see \Cref{fig:data}) but the SC dataset had significantly more positive responses with substantial length.For negative responses, we found that for WP, negative sentiment responses were only slightly longer than the positive ones, while there was a massive difference for SC. The mean length of SC negative sentiment responses is twice that of positive sentiment responses, and the 75\%-tile nearly doubles as well. This suggests that critical responses in the SC dataset are more informative than positive ones, and significantly more informative than ones in the WP dataset.

For toxicity analysis, we found detoxify's \texttt{toxicity} and \texttt{insult} measures to be the most useful way to quantify which critiques were non-constructive. We apply these measures to evaluate how toxic and insulting (respectively) negative responses are in each dataset. Scores given for these measures by the model are within the range of 0 and 1, 0 specifying the model does not think the text contains any toxicity or insults, and 1 specifying that the model is certain the text is toxic/insulting. When given to detoxify, SCs negative responses outputted near 0 \texttt{toxicity} and \texttt{insult} scores for the vast majority of responses.

\begin{table*}[ht]
\centering
\begin{tabular}{ccccc} 
Dataset & Measure & Est. Frequency \\
\midrule
\multirow{2}{*}{Writing Prompts} & Toxicity & 0.38 \\
 & Insulting & 0.17 \\
\vspace{-0.1cm}\\
\multirow{2}{*}{Story-Critique} & Toxicity & \textbf{0.15} \\
& Insulting & \textbf{0.03} \\
\end{tabular}
\caption{Estimated proportion of negative responses that are potentially toxic or insulting. To err on the side of caution, we identify a response as potentially toxic or insulting if the respective score is over a threshold of $0.01$.}
    \label{tab:toxicity}
\end{table*}

\section{Model Architecture and Training}\label{sec:model}

Our contrastive setup is largely identical to CLIP in that it contains two encoders, from which we compute an inner product of embeddings. Unlike CLIP, our model uses two text encoders; one for passages and one for critiques. For the text encoders, we use dual masked language models - similar to those used for DPR \citep{karpukhin2020dense}. Each encoder follows up the masked language model with an aggregate pooling function that performs a masked sum over the last layer of embeddings. This summed embedding is then normalized and fed to a fully connected layer that projects it to the encoding space (with dimensionality 2048). As with CLIP, we can then calculate cosine similarities between review and passage embeddings. However, we also multiply all of these similarities by a learned temperature. We clamped this temperature to fall in the interval $[\ln(1/100), \ln(100)]$ as we found this to stabilize training. We used a contrastive batch size of 2048 during training, which we chose using hyperparameter tuning.

We use RoBERTa \citep{liu2019roberta} as a base model for both encoders. We build three models, CARP-Tiny (58M params), CARP-Base (252M params), and CARP-Large (715M params), using RoBERTa-Tiny, RoBERTa-Base, and RoBERTa-Large for the encoders respectively. Note that our models possess two encoder branches, resulting in a total size approximately twice that of the corresponding RoBERTa model.

\begin{table*}[ht]
\centering
\begin{tabular}{lccccl} 
Model & Parameters (M) & Val Loss & Val Accuracy \\
\midrule
CARP-Tiny & 58 & 5.06 & 0.142 \\
CARP-Base & 252 & 4.76 & 0.148 \\
CARP-Large & 715 & \textbf{4.51} & \textbf{0.176} \\ 
\end{tabular}
    \caption{Different architectures and their performance after 1,400 steps. All use their respective variants of RoBERTa as text encoders.}
    \label{tab:archtable}
\end{table*}

The original CLIP pretraining objective uses cross-entropy to align embeddings of one modality to their corresponding embeddings in the other modality. We use a variant of this objective during training that first embeds both modalities without gradients, then slices over the batch to embed chunks and calculate loss per chunk, from which we accumulate gradients. Accumulating gradients over these chunks drastically reduces memory usage and allows the model to be trained with much larger contrastive batch sizes.

\begin{figure*}[ht]
{}\centering
    \includegraphics[scale=0.6]{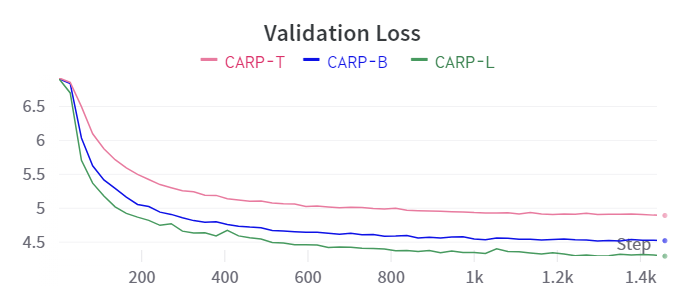}
    \vspace{0.7in}
    \includegraphics[scale=0.6]{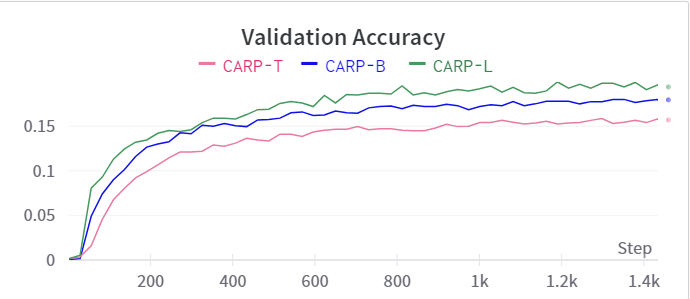}
\vspace{-0.75in}\caption{Validation Loss and Accuracy for CARP during training.}
\label{fig:training}
\end{figure*}

We hold out 1000 samples from the dataset as a validation set. We periodically compute loss and accuracy over this validation set for CARP-Tiny (CARP-T), CARP-Base (CARP-B), and CARP-Large (CARP-L) as shown in \Cref{fig:training}.

During inference we used Pegasus \citep{zhang2020pegasus} to perform prompt softening, paraphrasing our classifier four ways. We then report the average cosine similarity between an embedded story and these classifiers, effectively creating an ensemble of classifiers.

\section{Evaluation}\label{sec:evaluation}

We compare all three of our CARP models against three baseline models to see which more closely models the preferences of a human reader. 

\subsection{GPT-J Baseline Models}

We develop three baseline models to compare CARP to, all using GPT-J-6B \citep{gpt-j,gao2020pile}. For our first baseline model, we finetuned GPT-J-6B to generate critiques of passages in a seq2seq manner. Specifically, we concatenated strings of the form
\begin{verbatim}
    Passage: $PASSAGE
    Critique: $CRITIQUE
    <|endoftext|>
\end{verbatim}
to fill GPT-J-6B's 2048 token context window, where \texttt{\$PASSAGE} and \texttt{\$CRITIQUE} refer to an aligned passage and critic tuple.

For zero-shot baselines, we used two prompt engineered GPT-J without any fine tuning. For the first prompt engineered GPT-J, we used a prompting structure that mimicked the finetuning data. For the second prompt engineered GPT-J, we framed the prompt as a multiple choice quiz where a model needs to select which option is most likely given the passage.

To utilize this model for classification, we first write one review for each class by hand (see \Cref{sec:multichoice_prompts} for details). We then use the negative log likelihood per byte of each particular review given the passage as a proxy for classification score for each class. We compute this negative log likelihood for all classifiers, concatenate all these scalars into a vector, and normalize the resulting distribution. We use negative log likelihood per byte rather than per token to maintain tokenization agnostic.

To demonstrate the performance of this model, we report reviews generated by our seq2seq model, alongside the associated prompt input and perplexity in \Cref{tab:seq2seq}. The passages were crowdsourced from colleagues who were otherwise not involved in this project. We note that some generated reviews make specific references to the passage text, including cutting particular words and direct spelling or grammar revisions. 

\subsection{Experimental Design}

For each story, there are nine possible reviews that can be assigned to the story (all nine can be found in the appendix). These range from ``This kind of drags on'' to ``Could use more visual imagery.'' We encourage the human participants to select at least two reviews. On average, each participant selected 3.2 reviews. For every review for a story, we aggregate how many times it was voted for and use that as a score. We then normalize this score to obtain a distribution over the reviews. Normalization is performed by, for a given story, subtracting the minimum score from all other eight labels and then softmaxing.

We then provide our model with one of these stories, and compute the negative log likelihood (NLL) for each review. For the prompted models, this involves using perplexity as a surrogate for classification NLL, where as for \sysname{} we use cosine similarity as a surrogate for NLL. We similarly normalize these resulting distributions. Both the human baseline and the automated methods were normalized by first subtracting the minimum element and then softmaxing.

\subsection{Human Evaluation}

We recorded 200 participants over seven stories, where each participant received three stories. The stories were collected from colleagues unfamiliar with \sysname{}. We sourced stories since \sysname{} does not support sufficiently long stories as they go beyond the context length of the transformers comprising the text encoder. The same candidate labels were used for all stories. Both stories and labels can be found in the appendix. Participants were recruited via Prolific and paid \$16.50 USD/hour. We screened participants to select for college educated and English native speakers. Turkers were subjected to an attention screening question, where they were presented with an unambiguous story and asked to respond appropriately; we rejected 2\% of Turkers for answering this screening question incorrectly. The average completion time per Turker was 3.5 minutes.

\begin{figure*}[ht]
{}\centering
\includegraphics[width=\textwidth]{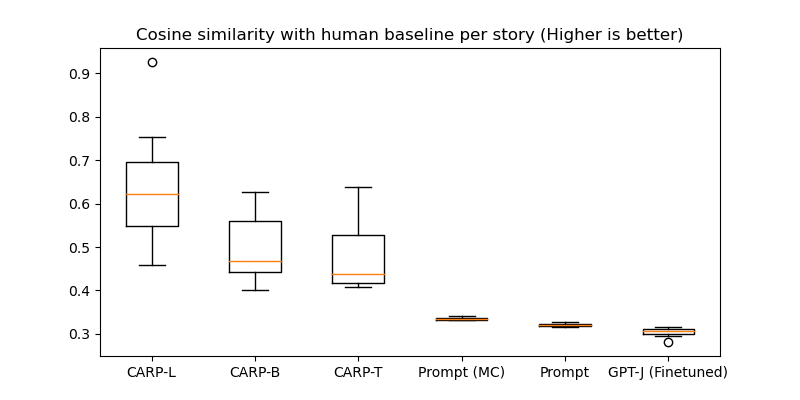}
\includegraphics[width=\textwidth]{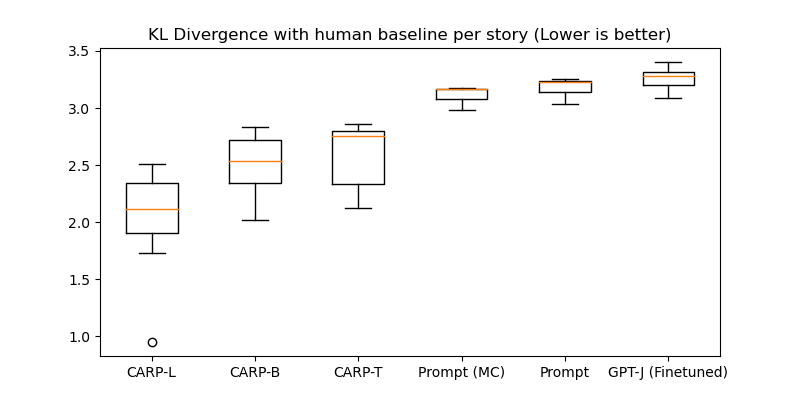}
\caption{We compare CARP-L, CARP-B, CARP-T, to our three baseline models: multiple choice classification prompt engineering, seq2seq classification prompt engineering, and seq2seq finetuning on our dataset. In the top plot we measure the cosine similarity (higher is better) of the predicted distribution against the human baseline. In the bottom plot, we compute $f(x) = \mathrm{KL}(\mathrm{softmax}(\mathrm{Human\ scores}), x)$ per story and similarly plot a box plot (lower is better).}
\label{fig:evaluation}
\end{figure*}

\section{Results}\label{sec:results}

As we can see in Figure \ref{fig:evaluation}, CARP performs strongly across the board when compared to other zero shot methods. Similarly, as we increase the size of CARP, performance improves linearly if not super linearly and, as a consequence, we can hypothesize that CARP benefits from scale. Finally, CARP costs a fraction of the cost of human evaluation, and as we can see correlates stronger with human evaluation when compared to zero shot methods. 

Notice that as we increase model size, we can see the mean performance for CARP improving. At the size of CARP-L, for example, the cosine similarity between the predicted distribution and the true distribution reaches 0.9 in some cases. We hypothesize that we are not near diminishing returns with CARP performance, and can still reap significant performance gains with increased scale.

Our finetuned GPT-J model has the worst performance out of all of the autoregressive models, and none of the autoregressive models perform anywhere near the human baseline on any of the tasks. This raises the question whether autoregressive models are suitable for zero-shot story evaluation, which we leave open for future work.


For STORIUM and Scarecrow, we note that STORIUM does not provide a human interpretable metric, so conducting a similar evaluation with STORIUM would not have been feasible. While Scarecrow is automated and human-interpretable, it is limited to only nine pre-determined evaluations and is not suitable for general use.

\section{Conclusions and Future Work}\label{sec:conclusion}

In this paper, we presented both the Story-Critique dataset as well as CARP. We believe that the methods outlined in CARP will result in moving towards story evaluation models that do not require expensive annotations and benefit massively from scale. 

In future work, we plan to extend this model to include larger contexts of windows of stories. Rather than being solely local critiques of stories, perhaps there is a method similar to \citet{khattab2021baleen} where one can perform critique as a multi-hop inference method, and accumulate critiques between hops. Similarly, expansions of the dataset to include both local and global critiques would be highly beneficial as long range critiques might prove to be a formidable NLU benchmark.

\ificlrfinal
\section*{Acknowledgments}
We would like to thank our anonymous data sources. We would also like to thank TRC (TPU Research Cloud) for providing us with the compute necessary to complete this project.

\fi
\bibliography{carp}
\bibliographystyle{iclr2022_conference}

\appendix
\ificlrfinal
\section{Author Contributions}
\textbf{Shahbuland Matiana:} Implemented and trained CARP, and worked on scaling.

\textbf{JR Smith:} Created and cleaned the dataset.

\textbf{Ryan Teehan:} Conducted data cleaning and obfuscation.

\textbf{Louis Castricato:} Conducted evaluation, designed prompt engineering, and oversaw the project.

\textbf{Stella Biderman:} Advised the design and analysis of the experiments, negotiated the use of the data, and led the writing of the paper.

\textbf{Leo Gao:} Trained the GPT-J baseline models.

\textbf{Spencer Frazier:} Editor, ethical oversight and assistance with dataset origination.

\fi

\section{CARP Prompting}

A major benefit of prompt engineering is that you can rapidly iterate over prompts without needing to curate new datasets or train new models. We found several ways of using prompt engineering on CARP to create zero shot classifiers.

Firstly, using emojis or emoticons allows you to build a sentiment classifier. A common occurrence within the dataset is the use of smiley faces within a review to signify that the editor liked the passage. Hence positive sentimental reviews act as a classification to signify that there are no major issues with the passage.

Secondly, editors often begin their critique with ``..." or a quote, signified in our dataset as a ``[quote]" token, when they wish to specify a syntactical flaw within the passage. Hence these act as a proxy for wording or grammatical errors.

Lastly, the critique ``Show, don't tell" acts as a good classifier for determining if a passage contains enough action. The issue, however, is that ``Show, don't tell" often requires some form of global information. It should be used sparingly.
\section{Baseline Model Prompting}

Below are the prompts utilized on the GPT-J-6B prompt engineered baseline. We utilized both an open-ended prompt and a multiple choice prompt.

\subsection{Sequence-to-Sequence Prompting}

The following prompt was used for ``Prompt" in the baselines:

\begin{displayquote}
Below is a set of stories and their critiques:

Passage: It was raining outside, so John got an umbrella out of the closet. He was off for an adventure.

Critique: Why does John want to go on an adventure, the transition is rather odd.

Passage: Jane wanted to go to the store to get cookies. She loved oatmeal cookies. 
Critique: Show, don't tell!

Passage: \$STORY
Critique: \$CRITIQUE
\end{displayquote}

\subsection{Multiple-Choice Prompting}\label{sec:multichoice_prompts}

The following prompt was used for ``Prompt (MC)" in the baselines. To improve readability, we replace newline characters with actual line-breaks.

\begin{displayquote}
You will be asked to read 4 stories and answer a set of questions about how you would classify the story. This first story is an example.\\
\\
Story: The rain was falling to the ground. \\
John took his umbrella off of the coat rack and walked out to a new adventure.\\
Choices:\\
A) This kind of drags on.\\
B) This is a bit too short.\\
C) This is too cheery.\\
D) This is really depressing.\\
E) This is really exciting.\\
F) This is boring.\\
G) This ending leaves things too open.\\
H) This ending feels abrupt.\\
I) Could use more visual imagery.\\
Answer: B)\\
\\
Question 1:\\
Story: Jane wanted to go to the store to get cookies. She loved oatmeal cookies. \\
Choices:\\
A) This kind of drags on.\\
B) This is a bit too short.\\
C) This is too cheery.\\
D) This is really depressing.\\
E) This is really exciting.\\
F) This is boring.\\
G) This ending leaves things too open.\\
H) This ending feels abrupt.\\
I) Could use more visual imagery.\\
Answer: F)\\
\\
Question 2:\\
Story: \$STORY\\
Choices:\\
A) This kind of drags on.\\
B) This is a bit too short.\\
C) This is too cheery.\\
D) This is really depressing.\\
E) This is really exciting.\\
F) This is boring.\\
G) This ending leaves things too open.\\
H) This ending feels abrupt.\\
I) Could use more visual imagery.\\
Answer: 
\end{displayquote}

\section{Sequence-to-Sequence Generated Critiques}\label{tab:seq2seq}

Below is an example of generated prompts using a seq2seq model finetuned on the Story-Critique dataset as described in earlier sections. We utilized top-p/top-k sampling (with hyper parameters of k=50 and p=0.9).

\begin{table*}[t]
    \centering
    \footnotesize
    \begin{tabular}{p{3.0in}p{1.2in}p{0.8in}}
    \textbf{Prompt} & \textbf{Critique} & \textbf{Perplexity}\\
    \midrule
    \textbf{Passage}: He grabbed the bag of flour, carefully measuring out a cup, making sure not to spill any on the counter. After pouring the flour into the bowl, he read over the recipe and noticed that he forgot to buy eggs, and wasn't able to make the cake. Rummaging through the fridge, he pulled out three eggs, which were still fresh in their carton. After cracking open one egg, he poured it into his mixing bowl along with two-thirds of a cup of sugar. He stirred them together until they combined smoothly, then added half a teaspoon of vanilla extract. \textbf{Critique:} & [quote]  I'm not sure what this means. & 1.265625\\
    
    \textbf{Passage}: Alice the goose went to the park because it was a nice day, especially for winter. She could see the bright sun reflecting off the thick ice on the pond. All the dogs were on leads so they didn't bother her, and being able to observe them from a safe distance let her appreciate their beauty and strength. After a nice warm swim in the pond, she got out and suddenly remembered she had many important goose duties to attend to. \textbf{Critique:} & [quote]  I don't understand this sentence. & 1.3369140625\\
    
    \textbf{Passage:} She climbed into her sports car and quickly drove down the coast to her favorite beach, enjoying the air conditioned interior on this hot day. She was looking forward to a nice cool swim after sweltering in the blazing heat all day. Upon arriving at the beach, she climbed off of her motorcycle and removed her helmet to shake out her hair.
    \textbf{Critique:} & [quote]  I think you can cut this. It doesn't really add anything to the story and slows down the pace. & 1.57421875\\
    
    \textbf{Passage:} John was going for a nice stroll on his local hiking path. It was snowing lightly. He loved the winter because it meant he didn't have to worry about any bugs bothering him, which was perfect since he'd had a strong fear of them since he was a child. He hated the buzzing and biting. This was exactly why John was hating this hike so far. The sweltering heat from the summer sun and the noisy buzzing of bugs made him very uncomfortable. \textbf{Critique:} & [quote]  I think you can cut this. & 1.3349609375\\
    
    \textbf{Passage:} The train ride to downtown Toronto was long and normally very crowded. Since it was very late at night, Jane did not see that many people entering or exiting the train, and the normally busy subway was empty and silent. Soon enough, the train had come to a complete stop. She got up and gathered her things as the doors opened.
    
    \textbf{Critique:} & [quote]  I think you can cut `very' from this sentence. It doesn't really add anything to the sentence. & 2.072265625\\
    
    \textbf{Passage:} Sam went down to the park to enjoy a nice walk in the sun. By the time he got there though, it suddenly started raining really heavily. His clothes were drenched and he was in a terrible mood. To make matters worse, when he got back home he discovered he'd left the stove on and his entire house had burnt down. \textbf{Critique:} & [quote]  I think you can cut this. It doesn't add anything to the story and it's a bit cliche. & 1.630859375\\
    
    \textbf{Passage:} It was a calm summers day. Birds were chirping, the sun was shining and the sky was clear of any clouds. George thought it was a wonderful day to go fishing. He pulled out on to the lake with his boat. \textbf{Critique:} & [quote]  I think you mean ``summer's day'' here, not ``summers day''. Also, I don't think you need to say that the sky was clear of clouds, since you already said that the birds were chirping and the sun was shining. & 1.849609375\\
\bottomrule
\end{tabular}
\end{table*}
\end{document}